\documentclass[letterpaper, 10 pt, conference]{ieeeconf}  %

\IEEEoverridecommandlockouts                              %

\usepackage[dvipsnames]{xcolor}

\usepackage{graphicx}
\usepackage{amsmath}
\usepackage{amssymb}
\usepackage{booktabs}
\usepackage{xcolor}%
\usepackage{subfiles}
\graphicspath{{imgs/}{../imgs/}}

\usepackage{bbm}

\usepackage{url}

\usepackage{wrapfig}
\usepackage{float}
\usepackage{tabu,multirow}
\restylefloat{table}
\usepackage{placeins}
\usepackage{booktabs}
\usepackage{footnote}
\usepackage{tablefootnote}
\usepackage{afterpage}
\usepackage{adjustbox}
\usepackage{arydshln}
\usepackage{gensymb}
\usepackage{caption}
\usepackage{cite}

\usepackage{mathtools}  

\usepackage[caption=false]{subfig} %
\usepackage{siunitx} %

\definecolor{cvprblue}{rgb}{0.21,0.49,0.74}
\usepackage[pagebackref,breaklinks,colorlinks,citecolor=cvprblue]{hyperref}
\usepackage{pgfplots}
\pgfplotsset{compat=1.9}
\pgfplotsset{
    myplotstyle/.style={
    legend style={draw=none, font=\small},
    legend cell align=left,
    legend pos=north east,
    ylabel style={align=center, font=\bfseries\boldmath},
    xlabel style={align=center, font=\bfseries\boldmath},
    x tick label style={font=\bfseries\boldmath},
    y tick label style={font=\bfseries\boldmath},
    scaled ticks=false,
    every axis plot/.append style={thick},
    },
}

\newcommand{\FullNETNAME}{IFFNeRF}
\newcommand{\NETNAME}{IFFNeRF }
\newcommand{\NETNAMEnoSpace}{IFFNeRF}

\title{\LARGE \bf \NETNAMEnoSpace: Initialisation Free and Fast 6DoF pose estimation \\ from a single image and a NeRF model}

\author{Matteo Bortolon$^{1,2,3}$, Theodore Tsesmelis$^{1}$, Stuart James$^{1,4}$, Fabio Poiesi$^{2}$ and Alessio {Del Bue}$^{1}$%
\thanks{*This work is part of the RePAIR project that has received funding from the European Union's Horizon 2020 research and innovation programme under grant agreement No.~964854.
This work has also received funding from the European Union’s Horizon Europe research and innovation programme under the project AI-PRISM (grant agreement No.~101058589)}%
\thanks{$^{1}$M. Bortolon, T. Tsesmelis, S. James, A. Del Bue are with Pattern Analysis and Computer Vision (PAVIS), Fondazione Istituto Italiano di Tecnologia (IIT), Genoa, Italy {\tt\small alessio.delbue@iit.it}}%
\thanks{$^{2}$M. Bortolon and F. Poiesi are with Technologies of Vision (TeV), Fondazione Bruno Kessler (FBK), Trento, Italy {\tt\small mbortolon@fbk.eu, poiesi@fbk.eu}} \thanks{$^{3}$M. Bortolon is with University of Trento, Trento, Italy} \thanks{$^{4}$S. James is with the Department of Computer Science, Durham University, UK {\tt\small stuart.a.james@durham.ac.uk}}%
}

\begin{document}

\maketitle
\thispagestyle{empty}
\pagestyle{empty}

\begin{abstract}
We introduce \NETNAME to estimate the six degrees-of-freedom (6DoF) camera pose of a given image, building on the Neural Radiance Fields (NeRF) formulation. 
\NETNAME is specifically designed to operate in real-time and eliminates the need for an initial pose guess that is proximate to the sought solution.
\NETNAME utilizes the Metropolis-Hasting algorithm to sample surface points from within the NeRF model. From these sampled points, we cast rays and deduce the color for each ray through pixel-level view synthesis.
The camera pose can then be estimated as the solution to a Least Squares problem by selecting correspondences between the query image and the resulting bundle.
We facilitate this process through a learned attention mechanism, bridging the query image embedding with the embedding of parameterized rays, thereby matching rays pertinent to the image.
Through synthetic and real evaluation settings, we show that our method can improve the angular and translation error accuracy by $\textbf{80.1\%}$ and $\textbf{67.3\%}$, respectively, compared to iNeRF while performing at 34fps on consumer hardware and not requiring the initial pose guess. Project page: \url{https://mbortolon97.github.io/iffnerf/}
\end{abstract}

\section{Introduction}\label{sec:intro}

Pose estimation is a cornerstone of robotic perception, that can also find application to autonomous vehicles and augmented reality.
The accuracy of this estimated pose is key to several downstream tasks, \textit{e.g.}~in robotics, it can influence the capability to manipulate objects and navigate within an environment \cite{marion2018, Manuelli2019kPAMKA, xu2023unidexgrasp, rajeev2019augmented}.
Recent advancements in neural and differentiable rendering, such as Neural Radiance Fields (NeRF) \cite{mildenhall2020nerf}, have unveiled innovative and efficient ways to model real-world scenes. 
NeRF-based models can train generative models that capture the structure and appearance of both objects and scenes.
They do that by leveraging images and their associated poses, eliminating the need for human supervision~\cite{mildenhall2020nerf} and unlocking novel view synthesis (NVS) in complex scenarios.
However, utilizing these models for subsequent camera re-localization in six degrees-of-freedom (6DoF) presents significant challenges. 
They typically demand an initial pose guess and rely on time-intensive optimization processes \cite{yen2020inerf, lin2023parallelinerf}.

Another benefit of NeRF-based approaches is that they can encode a broader range of appearance variations and eliminate the need for any mesh reconstruction \cite{bortolon2023vmnerf}.
iNeRF \cite{yen2020inerf} pioneered 6DoF pose estimation using NeRF through an analysis-by-synthesis method. 
Because NeRF can render an image of a novel view based on a specified pose, the problem can be inverted to deduce the pose from an uncalibrated image. 
This optimization process entails minimizing the photometric error between the supplied image and the ones rendered from potential poses, iterating until an error minimum is achieved.

\begin{figure}[t]
\centering
\includegraphics[width=1.\linewidth]{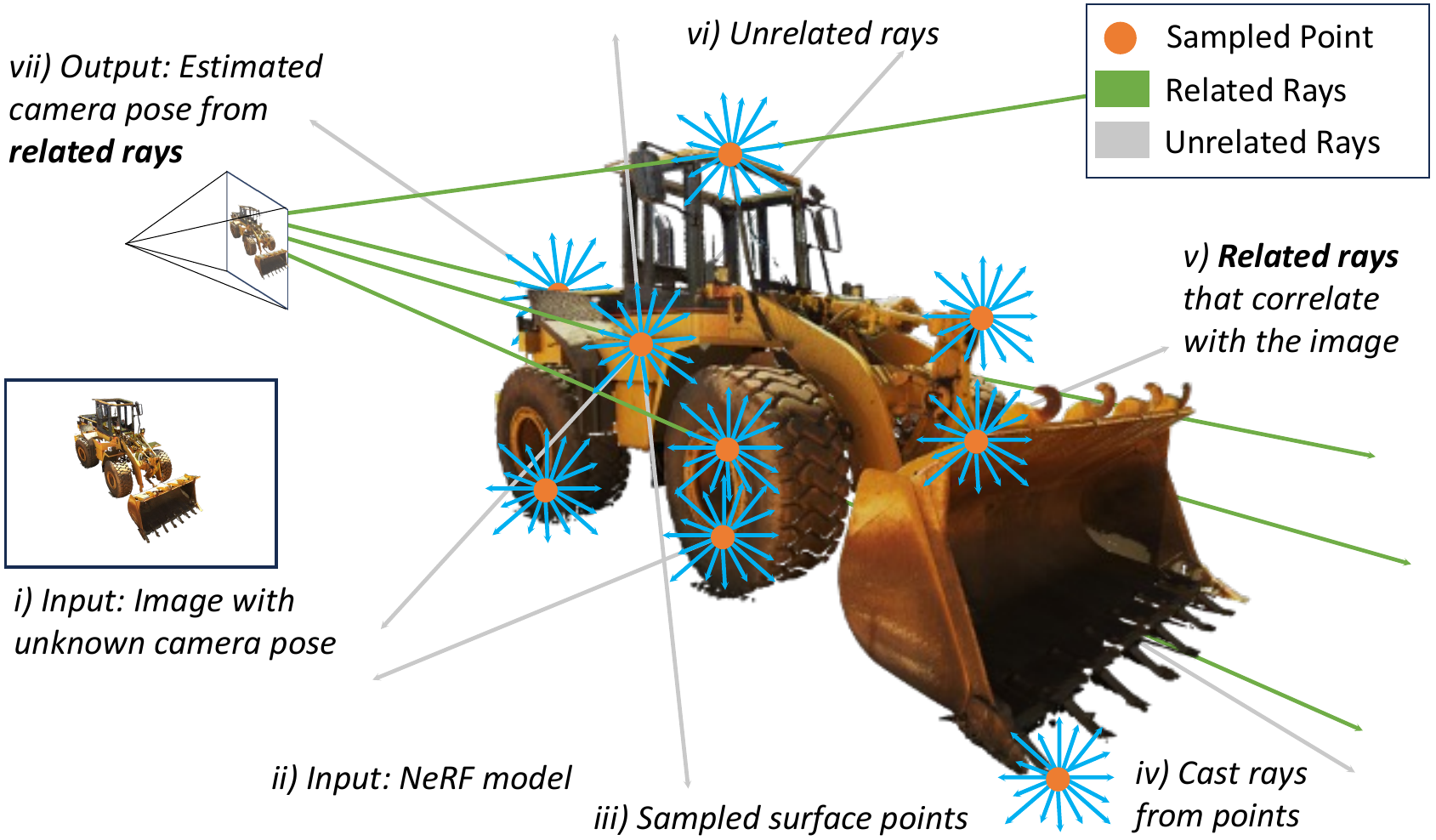} 
\vspace{-3mm}
\caption{From (\emph{i}) a given image with an unknown pose and (\emph{ii}) a NeRF model, we recover the pose by first (\emph{iii}) sampling surface points using Metropolis-Hasting algorithm and (\emph{iv}) casting rays from them in isocell distribution. 
We then (\emph{iv/v}) correlate rays with the image to identify relevant rays using attention and (\emph{vii}) recover the unknown 6DoF camera pose.}
\label{fig:camera_pose_recovery}
\end{figure}

To address the limitations of previous works, we introduce {\FullNETNAME}, an initialization-free method designed to estimate the 6DoF pose of a target image with respect to a NeRF model in real-time (Fig.~\ref{fig:camera_pose_recovery}).
Our approach employs surface points sampled using the Metropolis-Hastings (M-H) algorithm~\cite{metropolis1953equation} from within the volumetric NeRF model.
From these sampled points, we cast a series of rays that collectively span a wide range of potential views.
\begin{figure*}[t]
\begin{center}
\includegraphics[width=1\linewidth]{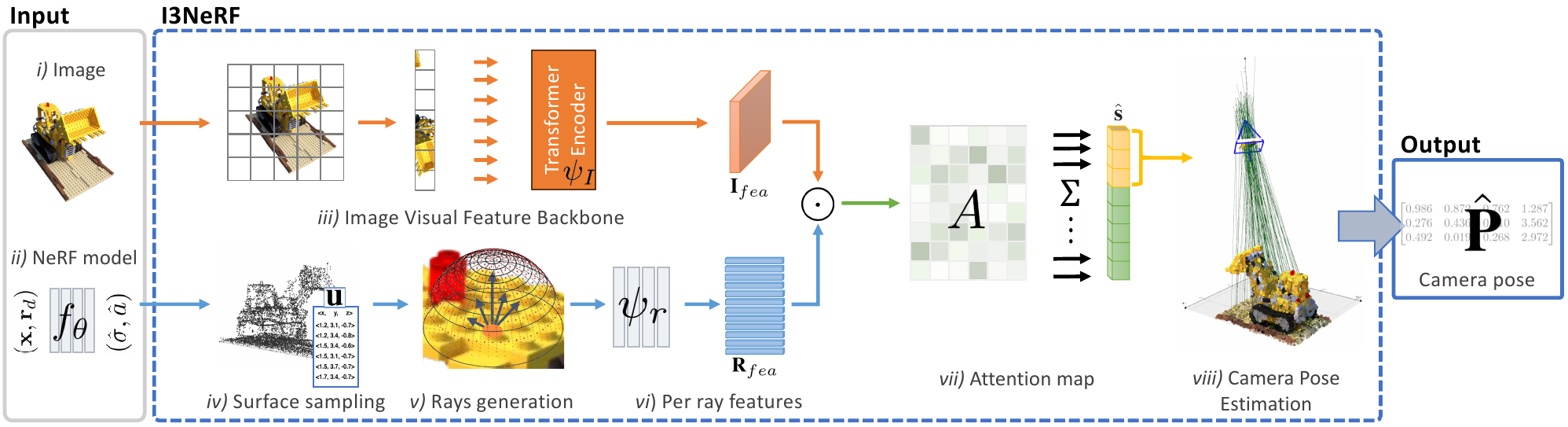 }
\end{center}
\vspace{-5pt}
\caption[]{Our \NETNAME takes as input an \emph{(i)} image, and \emph{(ii)} a NeRF model. The model $\psi_I$ encodes the image \emph{(iii)} using a visual backbone. As for the NeRF model, we sample surface points \emph{(iv)} using Metropolis-Hastings to identify candidates on the surface $\textbf{u}$, which can act as locations to project rays. We uniformly project rays \emph{(v)}  from the center isocells from the points, and estimate the ray corresponding parameters, color, and normal, using the NeRF model.  We then embed the ray representation $\psi_r$ \emph{(vi)}  and \emph{(vii)} learn attention $A$ between the ray embedding $\textbf{I}_{fea}$ and $\textbf{R}_{fea}$ to rank rays in relation to the image. We select the top-N rays and estimate the camera location \emph{(viii)} using least squares, resulting in a 6DoF pose $\hat{\textbf{P}}$ for the image.
}
\label{fig:teaser}
\end{figure*}
Each ray is defined by its origin, direction, and an associated surface color estimate, which is derived from pixel-level view synthesis.
To narrow down the vast array of potential rays, we employ a learned attention map between the query image embedding and the embedding of each cast ray~\cite{vaswani2017attention}, enabling us to identify a subset of rays that are relevant to the image.
This set of operations enables us to compute the target image pose in a closed-form way by solving a Least Squares problem.
We evaluate \NETNAME on both synthetic and real objects, benchmarking it against iNeRF.
Our experiments indicate that \NETNAME consistently outperforms its competitor, in particular when we search the camera pose in the wild (either no initial pose is given or is randomly initialized).
We further conduct an ablation study, scrutinizing different setups such as the attention map, backbone configurations, and various pose initializations.
Lastly, in contrast to iNeRF, we illustrate that \NETNAME operates in real-time and is more memory-efficient.
In summary, our contributions are:
\begin{itemize}
    \item Introducing a novel real-time method for estimating the 6DoF pose of a query image using a NeRF model;
    \item Eliminating the need for an initial camera pose, a requirement in iNeRF and similar methodologies;
    \item Proposing a new attention mechanism that proficiently evaluates the embeddings of cast rays and image pixels.
\end{itemize}

\section{Related work}
\label{sec:related_work}

The groundbreaking NeRF work demonstrated the potential of neural scene representations for creating novel photorealistic views~\cite{mildenhall2020nerf}.
Since then, the research community have significantly pushed the boundaries of neural radiance fields to improve Novel View Synthesis (NVS) accuracy and rendering/training performance, and apply NeRF to other tasks, including 6D pose estimation~\cite{yen2020inerf}.

Camera 6DoF pose estimation poses a critical perception challenge that can be tackled with deep learning or non-data driven methods~\cite{zhu2022review6DposeEstimation, marullo20236dreview}.
Among deep learning approaches, NeRF-based methods have recently gained relevant traction.
iNeRF~\cite{yen2020inerf} iteratively tries to aligns a query image and a rendered image by optimizing camera pose based on photometric error. 
This demonstrates accurate pose estimation in controlled environments like synthetic or static indoor scenes.
However, this accuracy comes with some shortcomings, including high inference time, inefficient pose updates, and dependence on accurate initial poses making it hard for iNeRF to perform outside of such controlled environments.

To address some of these issues, recent advancements have introduced parallel optimization using Monte-Carlo sampling~\cite{lin2023parallelinerf}.
In this case, instead of a single candidate pose, they generate and try to optimize multiple poses at the same time with different initial poses, and then prune and regenerate them based on an optimization trend.
This strategy improves convergence but does not eliminate the need for an initial camera pose while still manifesting high inference timings.
Similarly, Loc-NeRF employs Monte-Carlo and particle filtering to localize a robot within an environment using a 3D NeRF map~\cite{maggio2023locnerf}.
While this approach does not require prior pose information, it relies on multiple images and prior knowledge of their inter-image movement instead of a single image.

Alternative approaches, like CROSSFIRE~\cite{moreau2023crossfire}, incorporate learned descriptors into the NeRF model.
CROSSFIRE utilizes a joint training process of a visual backbone and the NeRF model to align 3D and 2D. During testing, CROSSFIRE adopts an analysis-by-synthesis approach, minimizing feature discrepancies rather than color differences.
However, both CROSSFIRE and parallel iNeRF need accurate initial camera positions to optimize for the actual camera pose.

While our method relies on a visual backbone to process the 2D image into features as CROSSFIRE, it does not follow an analysis-by-synthesis approach of a candidate image.
Instead we propose an approach that only synthesizes a small volume of the 3D model, allowing us to address some of the aforementioned shortcomings.

\section{Preliminaries}\label{sec:prelim}
\NETNAME exploits NeRF formulation~\cite{mildenhall2020nerf} whose objective is to synthesize novel views of a scene by optimizing a volumetric function \(f_{\mathbf{\theta}}\) from a given set of input images $\{\textbf{I}\}_{i=0}^{N-1}$ and corresponding camera poses $\{\textbf{P}\}_{i=0}^{N-1} \in \mathbb{R}^{3 \times 4}$. 
\(f_{\mathbf{\theta}}\) is parameterized through the weights of a Multi-Layer Perceptron (MLP).
The input to \(f_{\mathbf{\theta}}\) is defined as \((\hat{\sigma}, \hat{\mathbf{a}}) \leftarrow   f_{\mathbf{\theta}}(\mathbf{x}, \mathbf{r}_{d} )\), where $\mathbf{x}$ is a 3D point, $\mathbf{r}_{d}$ is a viewing direction, $\hat{\sigma}$ is a density, and $\hat{\mathbf{a}}$ is a color.
A ray marching method generates a set of rays starting from the camera's optical center and going through the scene. 
Rays are lines with a single point of origin $\mathbf{r}_{o}$ that extends infinitely in one direction $\mathbf{r}_{d}$.
We define a generic ray $\mathbf{r}(t) = \mathbf{r}_{o} + t \mathbf{r}_{d}$ where $t \in \mathbb{Z}$ is the position along ray.
The locations are defined between two clipping distances near ($t_n$) and far ($t_f$).
To make volumetric rendering computationally tractable, a finite set of 3D points is sampled along each ray $\{t_0, t_1, ..., t_{\Gamma-1}\}$, where \(\Gamma\) is the number of sampled locations.
NeRF employs a volumetric rendering function $\hat{\mathbf{c}}(\mathbf{r})$ to convert a ray into a color:
\begin{equation}
\label{eqn:rendering_formula}
\resizebox{.87\hsize}{!}{$
\hat{\mathbf{c}}\left( \mathbf{r},\Gamma \right) = \sum_{i=0}^{\Gamma - 1} s(i) \left ( 1 - e^{-\hat{\mathbf{\sigma}}(\mathbf{r}(t_i)) \mathbf{\delta}_i} \right ) \hat{\mathbf{a}}(\mathbf{r}(t_i)),$
}
\end{equation}
where $\mathbf{\delta}_i = t_{i+1} - t_i$ is the distance between adjacent sampled 3D points, and \(s(i)\) is the inverse of the volume density that is accumulated up to the \(i^{th}\) spatial location, which in turn is computed as
\begin{equation}\label{eqn:volume_density}
    \resizebox{.53\hsize}{!}{$
    s(i) = e^{ - \sum_{j=0}^{i-1} \hat{\mathbf{\sigma}}(\mathbf{r}(t_j)) \mathbf{\delta}_j},$
    }
\end{equation}
where \((1 - e^{- \hat{\mathbf{\sigma}}(\mathbf{r}(t_j)) \mathbf{\delta}_i})\) is a density-based weight component: the higher the density value \(\hat{\mathbf{\sigma}}\) of a point, the larger the contribution on the final rendered color. \(f_{\mathbf{\theta}}\) is therefore trained on a photometric loss minimizing $\mathcal{L} = \sum_{r} || \mathbf{c} - \hat{\mathbf{c}} ||^2_2$, where $\mathbf{c}$ is the true color of the pixel.
In the standard NeRF formulation, the model does not predict any volume points and their normals on the object's surface.
To obtain both the 3D points ($\mathbf{x}$) and their corresponding normals ($\mathbf{\hat{n}(x)}$), we adopt the Ref-NeRF approach~\cite{verbin2022refNerf}.

In more recent architectures \cite{chen2022tensorf, mueller2022instant}, the MLP is replaced by fast specialized data structures, these include hashmaps~\cite{mueller2022instant} and grids~\cite{chen2022tensorf}.
We take advantage of these speed improvements in our NeRF model by adopting the TensoRF work as our baseline~\cite{chen2022tensorf}.

\section{Our approach}\label{sec:method}

\NETNAME aims to predict the camera pose $\hat{\textbf{P}} \in \mathbb{R}^{3 \times 4}$ given an observed image $\textbf{I}$ and a pre-computed NeRF model $f_\theta$ (as define in Sec.~\ref{sec:prelim}). We firstly apply a Metropolis-Hastings algorithm to sample $G$ surface points within the scene volume (Sec.~\ref{sec:surface_sampling}), then we cast $V$ rays $\tilde{\mathbf{r}}$ from an isocell at each surface point (Sec.~\ref{sec:isocell_generation}), thus obtaining $\textbf{R}$ containing $G \times V$ rays. %
We then learn an attention map $A$ between embeddings of the image $\psi_I(\textbf{I})$ and generated rays $\psi_r(\mathbf{R})$ (Sec.~\ref{sec:attention_map}). 
Based on the information contained in the attention map, we select a subset of candidate rays that are likely to fall within the image. 
Finally, to recover $\hat{\textbf{P}}$ at test time, we optimize using Least Squares over the selected rays (Sec.~\ref{sec:pose_recovery}).

\subsection{Surface point sampling}\label{sec:surface_sampling}

We begin by extracting from the 3D points $\mathbf{x}$ a number $G$ of points $\mathbf{u}$ on the surface of the object.
To achieve this, we employ the Metropolis-Hastings (M-H) algorithm.
M-H works by sampling from a distribution $\mathcal{Q}$, in our case this distribution is the NeRF density output $\mathcal{Q} = \hat{\mathbf{\sigma}}$~\cite{metropolis1953equation}. 
We bootstrap the M-H algorithm by uniformly sampling points inside the bounding box that encloses the object's surface:
\begin{equation} \label{eq:surface_sampling}
    \mathbf{u}_0 = \boldsymbol{\beta} \otimes (\textbf{B}_{max} - \textbf{B}_{min}) + \mathbbm{1}_G \otimes \textbf{B}_{min},
\end{equation}
where $\mathbf{u}_0 \in \mathbb{R}^{G \times 3}$ are the sampled points at iteration $0$,
$\boldsymbol{\beta} = [\beta_0, \beta_1, ..., \beta_{G-1}]^T$ with each $\beta \sim \mathcal{U}[0, 1]$,
$\mathbbm{1}_G$ is a column vector of ones of length G,
$\otimes$ is the Kronecker product,
and $\textbf{B}_{max}$, $\textbf{B}_{min} \in \mathbb{R}^{1\times3}$ are the maximum and minimum 3D coordinates of the scene bounding box, respectively.

The M-H algorithm moves the point $g$, with  $g = 1 \ldots G$, only if $\mathcal{Q}(\mathbf{u}_{i+1,g}) \ge q$, where $q$ is the acceptance ratio. 
If the condition is satisfied,  $\mathbf{u}_{i+1,g}$ is a newly sampled random position.
We define $q$ as dependent from the Cumulative Distribution Function (CDF) $Pr$ of the previous iteration:
\begin{equation} \label{eq:alpha_surface_sampling}
q = Pr[Q_{i-1} < Q_{i-1,g}] \ge 0.6,
\end{equation}
that indicates $q$ as the $60^{th}$ percentile, of the density of the selected points at the previous iteration.
We define the points that are sampled at the last iteration of M-H as $\mathbf{u}$. 
These points are the ray origin, $\mathbf{\tilde{r}}_{o}$, for the following ray generation step, where we indicate with $\mathbf{\tilde{r}}$ the generated ray.

\subsection{Rays generation from sampled surface points}\label{sec:isocell_generation}

\begin{figure}[t]
\begin{center}
\subfloat[][{\footnotesize Isocell unit disc}]{\includegraphics[width=.30\linewidth]{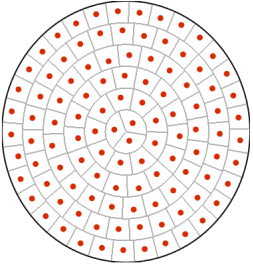}\label{fig:isocell}}
\hspace{0.01em}
\subfloat[][{\footnotesize Isocell unit hemi-sphere}]{\includegraphics[width=0.39\linewidth]{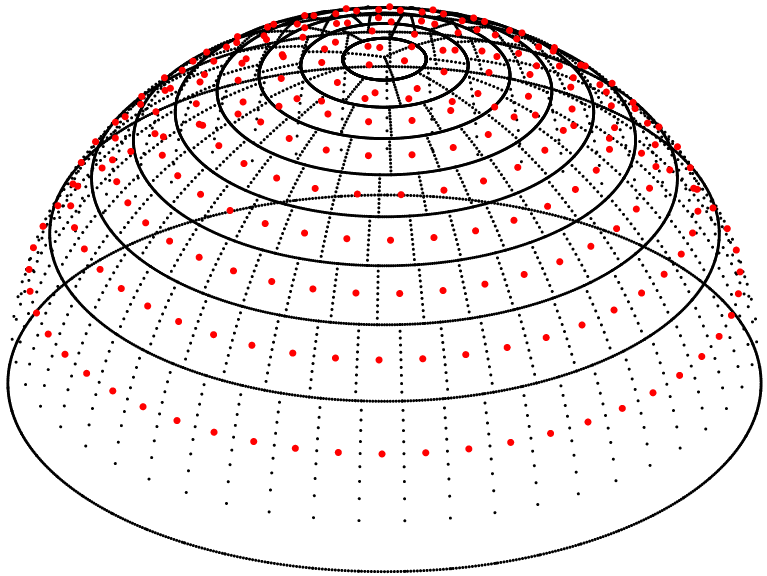}\label{fig:isocell_3d}}
\vspace{5pt}
\caption{Illustration of the Isocell ray generation method over a circular domain of a unit disk and unit sphere. 
The generated points indicate the ray positions within the equally spaced circle cells (we will denote them as ``cell centres'' for simplicity).}
\label{fig:ray_methods}
\end{center}
\vspace{-7pt}
\end{figure}

\begin{figure}[t]
\begin{center}
\includegraphics[width=.95\linewidth]{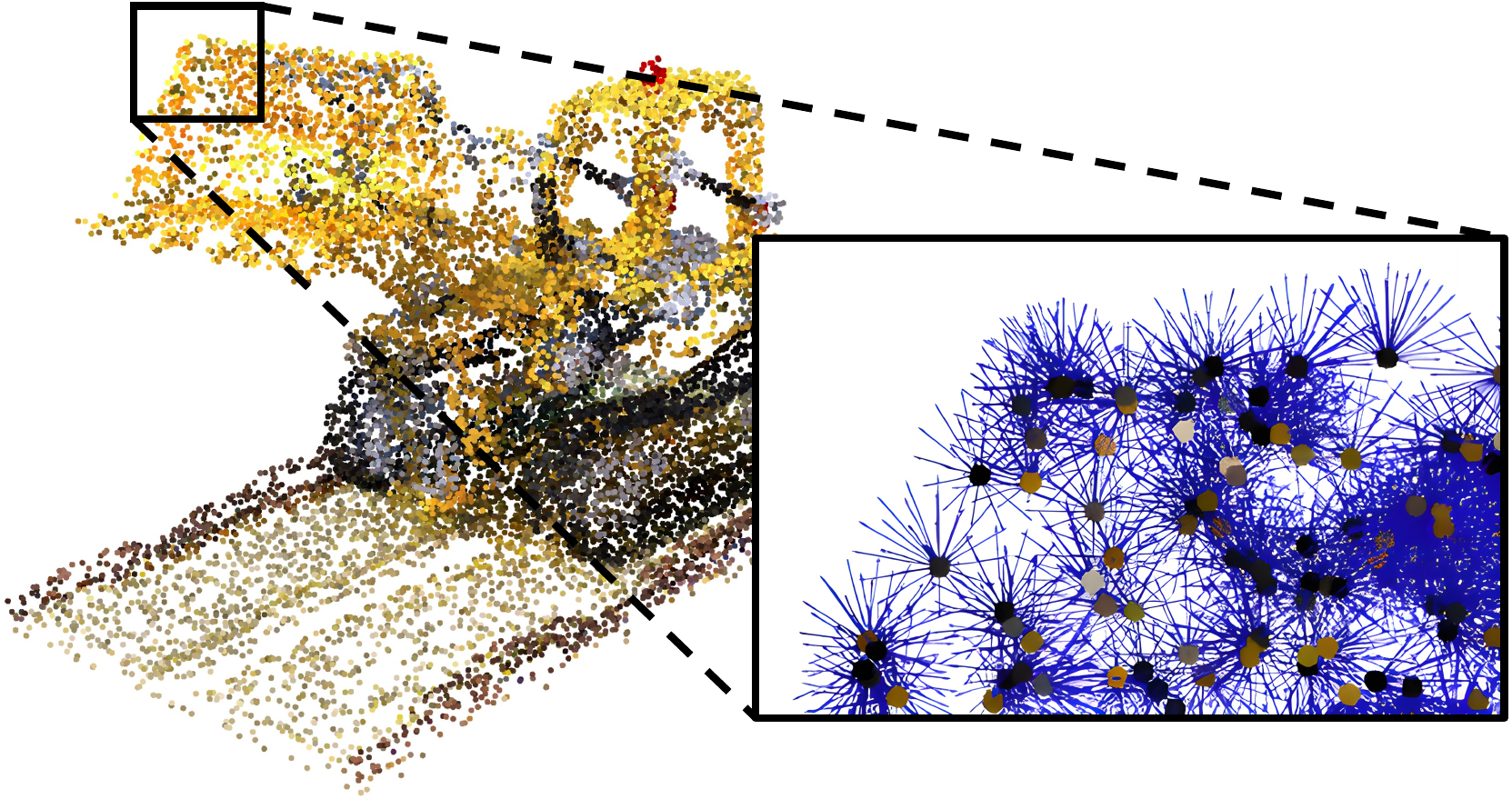}
\end{center}
\vspace{-5pt}
\caption{Example of generated rays using our approach.
The zoomed region highlights the ray cast operation (in this case $27$ rays per isocell).}
\label{fig:ray_sampling}
\end{figure}

From our sampled points $\mathbf{u}$, we cast a fixed number of rays in the direction oriented by the normal direction $\hat{\mathbf{n}}(\mathbf{u}_g)$, thus avoiding to cast rays toward the unobserved internal volume of the object.
The normals $\hat{\mathbf{n}}(\mathbf{u})$ are obtained by querying the NeRF model.
We adopt a deterministic method termed as Isocell \cite{masset2011, beckers2016}, which, compared to other commonly used rays sampling approaches, \textit{e.g.} Monte-Carlo \cite{malley1988}, achieves higher precision with fewer rays~\cite{jacques2015isocellEfficiency, Tsesmelis2018RGBD2luxDL}.
The uniform ray distribution of an isocell partitions the surface of a unit sphere into equal area cells, where the cells center forms a uniform distribution on the unit sphere (Fig.~\ref{fig:ray_methods}).
The distribution of an isocell provides the direction of the generated rays, identified as $\mathbf{\tilde{r}}_{d}$.
Therefore, we cast $V$ rays for each point $\mathbf{u}_g$, as shown in Fig.~\ref{fig:ray_sampling}.
We set $V=27$ as we empirically found that it works well in practice.

The generated rays represent a collection of potential hypotheses, meaning that a subset of them will intersect the target image $\textbf{I}$. 
In total, we have $N = V \times G$ generated rays $\tilde{\mathbf{r}}$ across the sampled points.
Each generated ray have also a color $\mathbf{\tilde{r}}_{c}$, that it is computed through the same pixel-level approach of NeRF (Eq.~\ref{eqn:rendering_formula}). 
Note that, the application of the volumetric rendering function of Eq.~\ref{eqn:rendering_formula} produces a single pixel for each ray.
Moreover, if we assume an object is made of opaque surfaces, we can make our algorithm computationally more efficient by sampling only a few points located near the ray's origin (points that that are further away from the origin does not influence the rendered color).

In summary, the generated rays are defined by their origin $\mathbf{\tilde{r}}_{o}$, direction $\mathbf{\tilde{r}}_{d}$, and rendered color $\mathbf{\tilde{r}}_{c}$.
To simplify the notation, we group and term these three components as $\tilde{\mathbf{r}}$.
We can now leverage all the ray features to associate the rays with the image pixels.

\subsection{Matching by attenuating cast rays to image}\label{sec:attention_map}

From all the cast rays $\tilde{\mathbf{r}}$, we identify a subset of them that maximally correlates with the given image $\textbf{I}$ in order to recover the image pose ($\hat{\textbf{P}}$). 
To avoid a time-consuming brute force search, we learn to score rays that are related to the image pixels. 
We achieve this by embedding ray parameters and the image into a higher-dimensional feature space.

Specifically, to enable our approach to discriminate rays that share similar appearance and position, we utilize a multi-layer perceptron (MLP) denoted as $\textbf{R}_{fea} = \psi(\mathbf{\tilde{r}})$, where $\textbf{R}_{fea} \in \mathbb{R}^{N \times C}$, and $C$ denotes the number of channels.
This MLP processes each ray independently, making the procedure insensitive to changes regardless how the rays are ordered.
We project the MLP input into a higher-dimensional space using NeRF positional encoding~\cite{mildenhall2020nerf}.
This positional encoding expands subtle differences in the input, enabling the network to better distinguish similar rays~\cite{tancik2020fourfeat}.

To represent $\textbf{I}$, we use DINOv2~\cite{oquab2023dinov2} feature extractor, resulting in a set of features $\textbf{I}_{fea} \in \mathbb{R}^{WH \times C}$, where $W$ and $H$ are the width and height of the image backbone's output, respectively.
Then, we define an attention module, $\textbf{M} \in \mathbb{R}^{W \times H \times C}$ to allow our model to correlate ray features, $\textbf{R}_{fea}$, and image features, $\textbf{I}_{fea}$, by using the image feature as query and the ray features as key: $\textbf{M} = A(\mathbf{R}_{fea}, \mathbf{I}_{fea})$.

We optimize the attention map by summing along the rows and converting it into a correlation score, $\hat{\textbf{s}}$ per ray as:
\begin{equation}\label{eqn:pred_correlation_score}
    \resizebox{.35\hsize}{!}{$
    \hat{\textbf{s}}= \sum_{k=1}^{WH} \textbf{M}_{k}.$
    }
\end{equation}
The ground-truth score $\textbf{s}$ is defined based on the distance between the camera and its projection onto the considered ray.
We can compute the projection of the point on the line with:  $t = max((\textbf{p} - \tilde{\mathbf{r}}_{o}) \tilde{\mathbf{r}}_{d}, 0)$,
where $\textbf{p}$ is the camera position, $\tilde{\mathbf{r}}_{o}$ the generated ray and $\tilde{\mathbf{r}}_{d}$ the corresponding direction.
Rays are infinite only in one direction, so we restrict $t \in \mathbb{R}^{+}$ using the max operator.

Then we can compute the distance between the camera origin and its projection on the ray as follows $\mathbf{d} = \|(\tilde{\mathbf{r}}_{o} + t  \tilde{\mathbf{r}}_{d}) - \textbf{p}\|_2$.
$\mathbf{d}$ is normalize in $[0, 1]$ with:
\begin{equation}
\label{eqn:score_regularization}
    \mathbf{d}_{\tilde{r}} = 1 - tanh\left(\frac{\mathbf{d}}{\lambda}\right),
\end{equation}
where $\lambda$ is a regularization parameter that adjusts the score values range.
This affects how the rays are correlated and ranked based on the image features.

Lastly, we normalize all the scores $\textbf{s}$ as follows:
\begin{equation}\label{eqn:score_normalization}
\textbf{s} = \mathbf{d}_{\mathbf{\tilde{r}}} \frac{HW}{\sum^{HW}_{j=0} \textbf{d}_{\mathbf{\tilde{r}, j}}}.
\end{equation}
During training our model optimizes $\hat{\textbf{s}}$ based on $\textbf{s}$ by using the $L2$ norm loss function.

\subsection{Test-time pose estimation}
\label{sec:pose_recovery}

\begin{figure}[t]
\begin{center}
\subfloat[][{\footnotesize Sampled rays}]{\includegraphics[width=.49\linewidth]{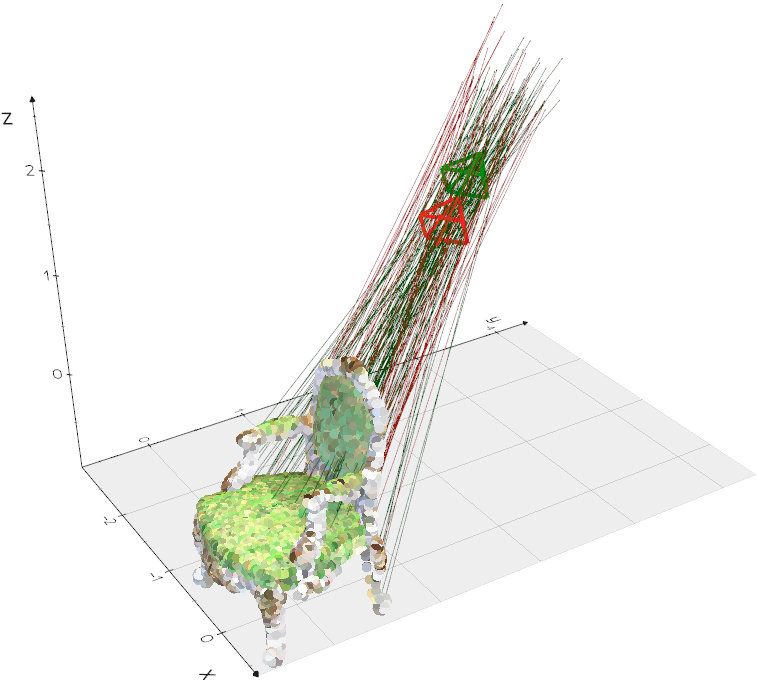}\label{fig:gt_est_rays}}
\hspace{0.05em}
\subfloat[][{\footnotesize Score distributions}]{\includegraphics[width=.49\linewidth]{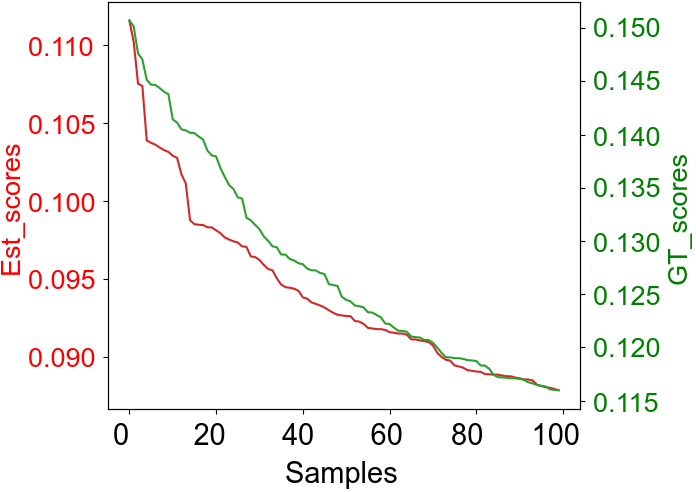}\label{fig:scores_distribution}}
\caption{Example of the test-time pose estimation on the Chair object of Synthetic NeRF.
(a) Top $N$ (red) \textit{vs.} top $N$ ground-truth (green) rays. 
(b) Corresponding score distribution of top $N$ and ground-truth rays.}
\label{fig:camera_pose_estimation}
\end{center}
\vspace{-7pt}
\end{figure}

At test-time the predicted scores $\hat{\textbf{s}}$, are used as a correlation metric to select the best $N_{top}$ rays, which should be the most relevant rays to the image, thus facilitating pose estimation.
Fig.~\ref{fig:camera_pose_estimation} shows an example of the selected rays and the score distribution. 
Note that only a small set of rays is sufficient to estimate the camera pose. 
However, we empirically set $N_{top}=100$ for robustness (Fig.~\ref{fig:gt_est_rays}).  

Because the camera pose estimation can be seen as the intersection point between the selected rays, we compute the ray intersection by searching for the solution of a weighted linear system of equations. Naturally, a set of 3D lines will not intersect at a single point, especially considering the noise introduced by discretizing the rays directions in the isocell.
To overcome this, we compute the Least Squares solution that minimizes the sum of the squared perpendicular distances.

Formally, for the generated ray, $\mathbf{\tilde{r}}_j$ with $j = 1 \ldots N_{top}$, the error is given by the square of the distance from point $\mathbf{p}$ to its projection on $\mathbf{\tilde{r}}_j$:
\begin{equation}\label{eqn:minimization_of_square_diff}
\resizebox{.88\hsize}{!}{$
    \sum_{j=0}^{N_{top} - 1}\left( (\mathbf{p}-\mathbf{\tilde{r}}_{o,j})^T (\mathbf{p}-\mathbf{\tilde{r}}_{o,j}) - ((\mathbf{p}-\mathbf{\tilde{r}}_{o,j})^T\mathbf{\tilde{r}}_{d,j})^2 \right).
        $}
\end{equation}
An efficient way to compute the least square solution to Eq.~\ref{eqn:minimization_of_square_diff} is to compute its derivative respect to $\textbf{p}$, resulting in
\begin{equation}\label{eqn:minimization_of_square_diff_differentiate}
    \resizebox{.7\hsize}{!}{$
    \mathbf{p} = \sum_{j=0}^{N_{top} - 1}\hat{\mathbf{s}}_{j}(\mathbb{I} - \mathbf{\tilde{r}}_{d,j}\mathbf{\tilde{r}}_{d,j}^T)\mathbf{\tilde{r}}_{o,j},$
    }
\end{equation}
where $\mathbb{I}$ is the identity matrix and $\hat{\mathbf{s}}_{i}$ the predicted ray scores.

\newcolumntype{"}{@{\hskip\tabcolsep\vrule width 1.3pt\hskip\tabcolsep}}
\begin{table*}[ht]
\centering
\begin{adjustbox}{max width=\textwidth}
\setlength\tabcolsep{1.5pt}
\renewcommand{\arraystretch}{1.3}
\begin{tabular}{l|c|c|c|c|c|c|c|c|c|c|c|c|c|c|c|c"c|c|r}
\multicolumn{20}{c}{(a) Synthetic NeRF}                                                                                                                                                                                                                                                                               \\
                                                  & \multicolumn{2}{c|}{Drums} & \multicolumn{2}{c|}{Chair} & \multicolumn{2}{c|}{Lego} & \multicolumn{2}{c|}{Ship} & \multicolumn{2}{c|}{Ficus} & \multicolumn{2}{c|}{Mic} & \multicolumn{2}{c|}{Hot Dog} & \multicolumn{2}{c"}{Materials} & \multicolumn{3}{c}{Mean}        \\
Method                                            & MAE         & MTE         & MAE         & MTE         & MAE        & MTE         & MAE        & MTE         & MAE         & MTE         & MAE        & MTE        & MAE          & MTE          & MAE           & MTE           & MAE   & MTE    & Time (s) \\ \hline
iNeRF (initialization by \cite{yen2020inerf}) & 17.4        & 0.430       & 15.4        & 0.441       & 12.7       & 0.293       & 9.8        & 0.344       & 18.1        & 0.485       & 17.6       & 0.501      & 3.8          & 0.061        & 15.1          & 0.375         & 13,7 & 0.366 & 6.1          \\ \hline \hline
iNeRF (Random initialization)                     & 74.2        & 1.493       & 100.5       & 2.571       & 93.0       & 1.868       & 89.1       & 1.535       & 93.9        & 1.985       & 99.3       & 2.520      & 86.5         & 1.707        & 84.3          & 1.734         & 90.1 & 1.927 & 6.1          \\
Ours (w. DINOv2 pretrained)                 & \textbf{22.5}        & 1.210       & \underline{13.2}        & 1.106       & \textbf{11.8}       & 1.009       & 50.5       & 1.318       & \underline{19.4}        & 1.417       & \textbf{10.7}       & 1.046      & \textbf{10.5}         & 1.164        & 33.1          & 1.252         & \underline{21.5} & 1.190 & \textbf{0.029}          \\
Ours (w. DINOv2 fine-tuned)                          & 28.3        & \textbf{0.916}       & \textbf{7.1}         & \underline{0.429}       & \underline{17.6}       & \underline{0.547}       & \underline{25.7}       & \underline{0.600}       & \textbf{18.8}        & \underline{0.818}       & 14.8       & \underline{0.576}      & 15.6         & \underline{0.650}        & \textbf{15.4}          & \underline{0.493}         & \textbf{17.9} & \underline{0.629} & \textbf{0.029}          \\ \hdashline
Ours (w. DINOv2 fine-tuned) w/iNeRF                  & \underline{26.2}        & \underline{0.921}       & 25.7        & \textbf{0.351}       & 18.4       & \textbf{0.462}       & \textbf{23.1}       & \textbf{0.524}       & 27.8        & \textbf{0.639}       & \underline{13.5}       & \textbf{0.521}      & \underline{15.4}         & \textbf{0.537}        & \underline{23.4}          & \textbf{0.455}         & 21.7 & \textbf{0.551} & 6.2
\end{tabular}
\end{adjustbox}
\vspace{10pt}

\resizebox{.75\textwidth}{!}{
\setlength\tabcolsep{1.5pt}
\renewcommand{\arraystretch}{1.3}
\begin{tabular}{l|l|l|l|l|l|l|l|l|l|l"l|l|r}
\multicolumn{14}{c}{(b) Tanks And Temples (real)}                                                                                                                                                                                               \\
                                                  & \multicolumn{2}{c|}{Barn} & \multicolumn{2}{c|}{Caterpillar} & \multicolumn{2}{c|}{Family} & \multicolumn{2}{c|}{Ignatius} & \multicolumn{2}{c"}{Truck} & \multicolumn{3}{c}{Mean}        \\
Method                                            & MAE        & MTE         & MAE            & MTE            & MAE         & MTE          & MAE          & MTE           & MAE         & MTE         & MAE   & MTE    & Time (s) \\ \hline
iNeRF (initialization by \cite{yen2020inerf}) & 26,5       & 0,208       & 42,9           & 0,166          & 42,8        & 0,794        & 31,4         & 0,723         & 31,6        & 0,370       & 35,0 & 0,452 & 6.1  \\ \hline \hline
iNeRF (Random initialization)                 & 89.2       & 0.682       & 89.3           & 2.559          & 93.9        & 1.505        & 84.1         & 1.489         & 94.4        & 1.042       & 90.2   & 1.455  & 6.1  \\
Ours (w. DINOv2 pretrained)             & \textbf{22.3}       & 0.121       & 33.2           & 0.472          & 24.3        & 0.638        & \textbf{15.6}         & 0.495         & 22.3        & 0.140       & \underline{23.5}   & 0.373  & \textbf{0.029}  \\
Ours (w. DINOv2 fine-tuned)                      & \underline{26.9}       & \underline{0.060}       & \underline{30.4}           & \underline{0.427}          & \underline{22.1}        & \underline{0.585}        &\underline{18.7}         & \underline{0.455}         & \textbf{20.7}        & \underline{0.104}       & 23.7   & \underline{0.326}  & \textbf{0.029}  \\ \hdashline
Ours (DINOv2 fine-tuned) w/iNeRF              & \underline{26.9}       & \textbf{0.037}       & \textbf{28.3}           & \textbf{0.395}          & \textbf{18.1}        & \textbf{0.561}        & 19.5         & \textbf{0.297}         & \underline{21.9}        & \textbf{0.096}       & \textbf{22.9}   & \textbf{0.277}  & 6.2 
\end{tabular}
}
\vspace{1mm}
\caption{Evaluation of 6DoF pose estimation on two datasets: (a) Synthetic NeRF \cite{mildenhall2020nerf} and (b) Tanks \& Temples \cite{Knapitsch2017TanksAndTemples}. 
We report the results in terms of Mean Angular Error (MAE) and Mean Translation Error (MTE).
The lower MAE and MTE, the better.
Best-performing results are highlighted in \textbf{bold} and second-best results are \underline{underlined}.
}
\label{tab:per_object_results}
\end{table*}

\section{Evaluation}
\label{sec:eval}
We compare \NETNAME to the baseline method iNeRF, with same backbone as ours\footnote{Parallel iNeRF and CROSSFIRE~\cite{lin2023parallelinerf, moreau2023crossfire} code and evaluation protocol were not available at the time of submission. Evaluating our method on their testing data was also impractical as testing samples in~\cite{lin2023parallelinerf} are sampled randomly from larger dataset and the samples information is not available either.}.
Using iNeRF's protocol, we compare results on Synthetic NeRF~\cite{mildenhall2020nerf} and, the real-world dataset, Tanks \& Temples~\cite{Knapitsch2017TanksAndTemples}. 
For each dataset, we use the predefined training-test splits and evaluate iNeRF with two initializations: 
\emph{i)} the initialization proposed by iNeRF, where a given starting pose is between $[-40^{\circ}, +40^{\circ}]$ degrees and of $[-0.2, +0.2]$ units of rotational and translation error from the ground-truth target pose, respectively; 
\emph{ii)} the initialization obtained by randomly selecting a pose from the ones available in the training set.
Additionally, we conduct an ablation study on \NETNAMEnoSpace's backbone configuration (pretrained \textit{vs.} fine-tuned) and the use of \NETNAME as an initialization for iNeRF.
Evaluation metrics include mean angular (MAE) and translation (MTE) errors, and we measure the inference time in Tab.~\ref{tab:per_object_results}.

\noindent\textit{Implementation Details:} 
We implement \NETNAME with PyTorch and train for 1.5K iterations ($\sim$45mins) on the training split on a standard workstation with a NVIDIA GeForce RTX 3090. 
We use the Adam optimizer with learning rate $10^{-3}$.
Hyper-parameters are set as $\lambda=1$, $G = 5000$ and $800$ M-H iterations.

\subsection{Datasets}

We use two datasets:
\emph{i) Synthetic NeRF} \cite{mildenhall2020nerf}: This dataset was released together with NeRF. 
Synthetic NeRF comprises eight scenes (Chair, Drums, Ficus, Lego, Materials, Ship, Mic, Hot Dog), includes both camera intrinsic and extrinsic parameters, provides train and test splits of $100$ images ($33\%$) and $200$ images ($66\%$), respectively.
\emph{ii) Tanks \& Temples}~\cite{Knapitsch2017TanksAndTemples}: This dataset was originally created to evaluate 3D reconstruction methods and embeds challenging scenarios as real-world objects with variations in size (small and large), acquired from a human-like viewpoints, at various distances from the object, and with different illumination conditions, such as shadows.
As in~\cite{chen2022tensorf}, we evaluate on five scenes: Barn, Caterpillar, Family, Ignatius and Truck.
We use the train and test splits as specified in~\cite{liu2020nsvf}, with the split depending on the object, having on average $\approx247$ training images ($87\%$) and $\approx35$ testing images ($12\%$).
Like~\cite{lin2023parallelinerf}, we resize all the objects for both datasets to fit inside a unit box. Therefore, the translation error is relative to the object size and we define it in units.

\begin{figure*}[t!]
\begin{center}
        \subfloat[][{\footnotesize Chair}]{\includegraphics[width=0.20\linewidth]{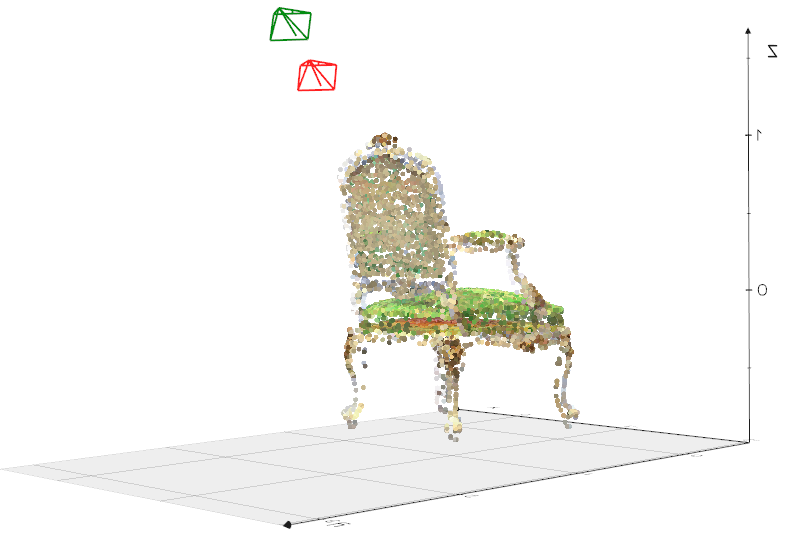}\label{fig:chair}}
		\hspace{0.01em}
		\subfloat[][{\footnotesize Bulldozer}]{\includegraphics[width=.20\linewidth]{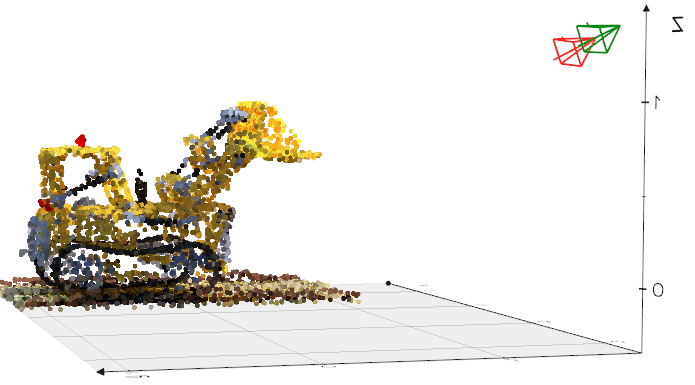}\label{fig:buldozer}}
		\hspace{0.01em}
        \subfloat[][{\footnotesize Microphone}]{\includegraphics[width=0.18\linewidth]{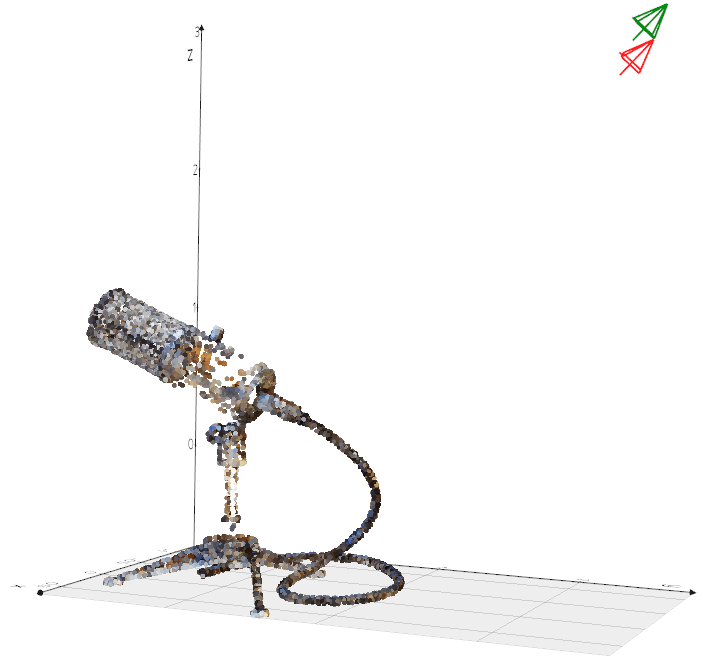}\label{fig:mic}}
		\hspace{0.01em}
		\subfloat[][{\footnotesize Hotdog}]{\includegraphics[width=.18\linewidth]{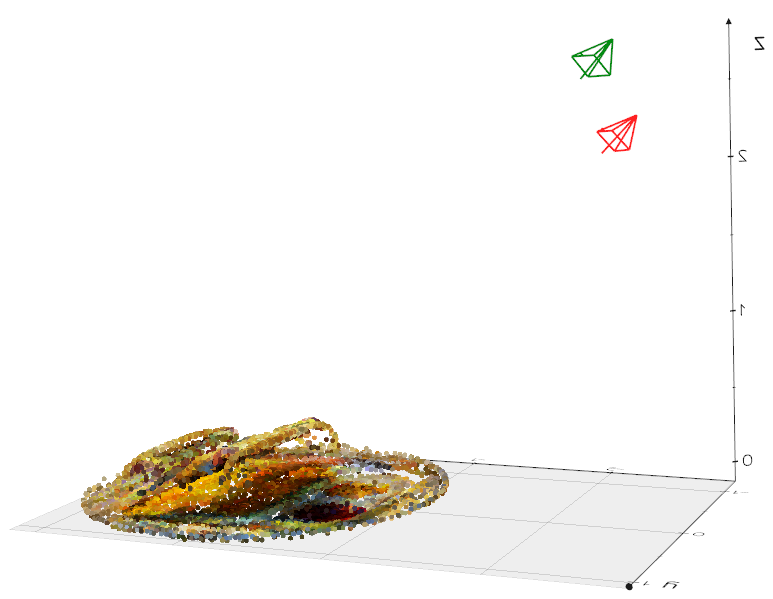}\label{fig:hotdog}}
		\hspace{0.01em}
		\subfloat[][{\footnotesize Materials}]{\includegraphics[width=0.18\linewidth]{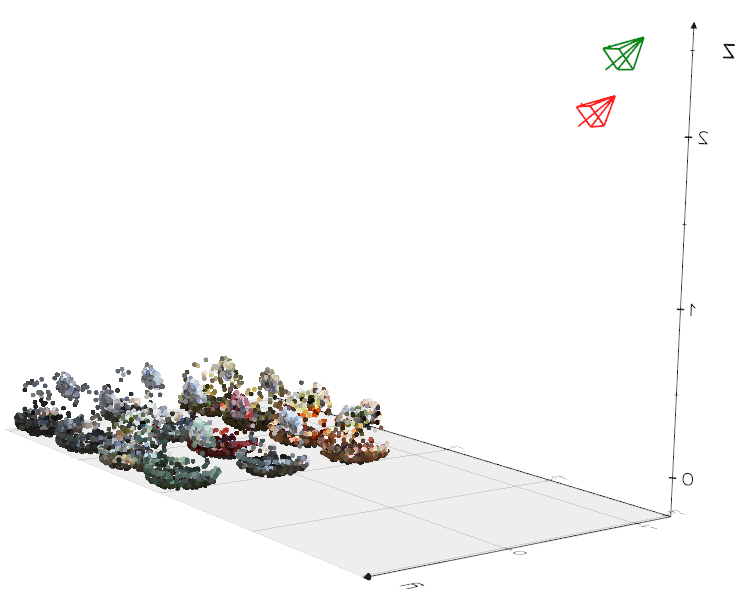}\label{fig:materials}}
        \\
        \subfloat[][{\footnotesize Barn}]{\includegraphics[width=0.23\linewidth]{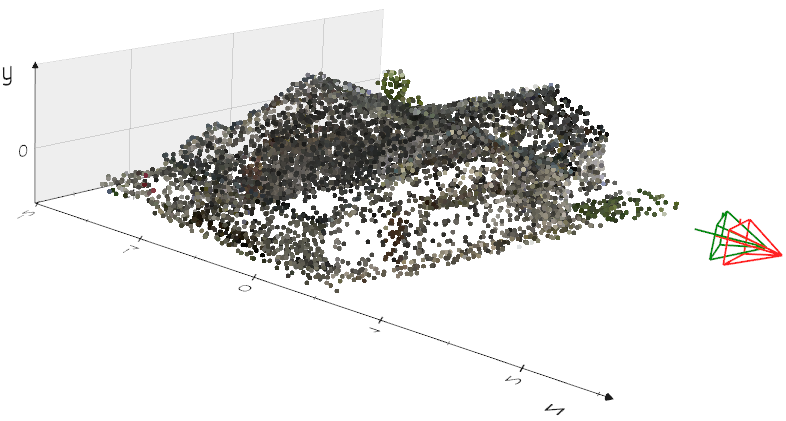}\label{fig:burn}}
		\hspace{0.01em}
		\subfloat[][{\footnotesize Truck}]{\includegraphics[width=.23\linewidth]{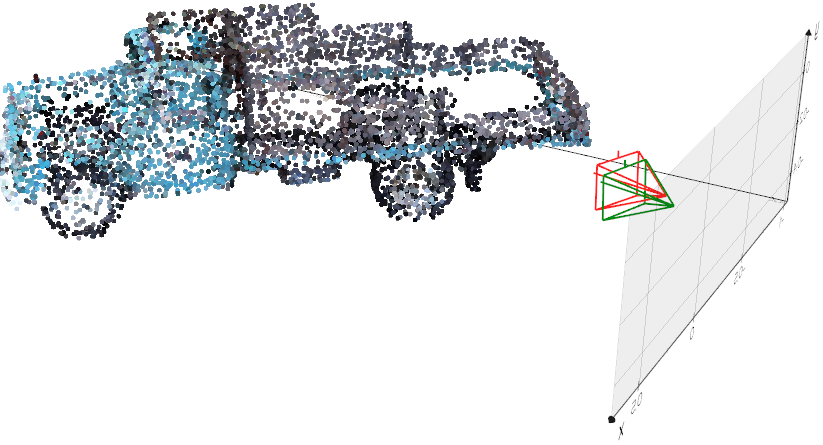}\label{fig:truck}}
		\hspace{0.01em}
        \subfloat[][{\footnotesize Caterpillar}]{\includegraphics[width=0.18\linewidth]{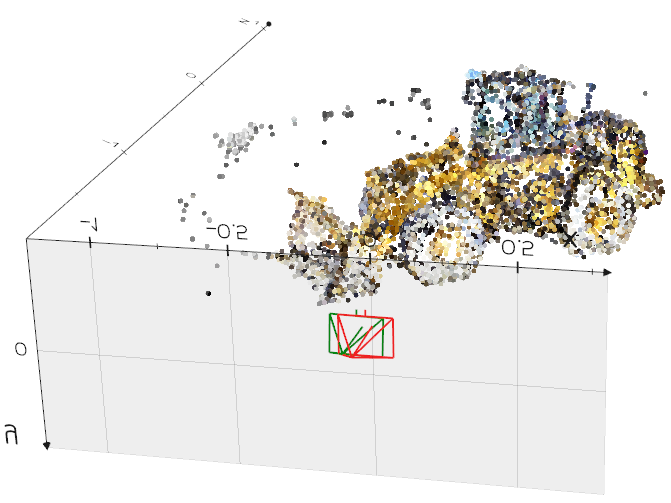}\label{fig:caterpillar}}
		\hspace{0.01em}
		\subfloat[][{\footnotesize Family}]{\includegraphics[width=.14\linewidth]{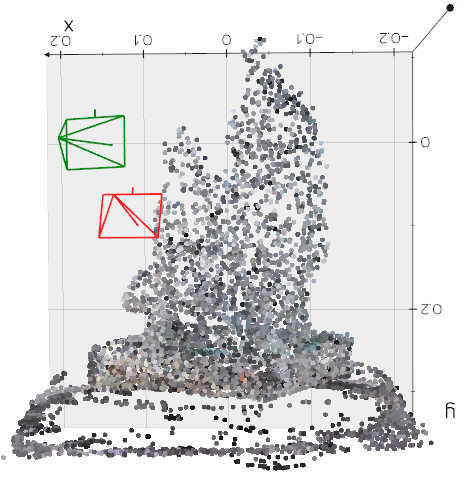}\label{fig:family}}
		\hspace{0.01em}
		\subfloat[][{\footnotesize Ignatius}]{\includegraphics[width=0.16\linewidth]{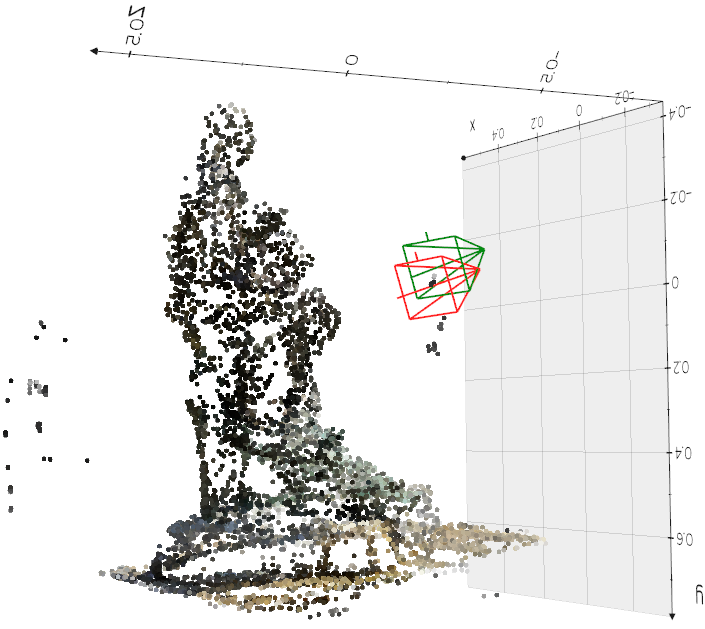}\label{fig:ignatius}}
        \vspace{5pt}
\end{center}
\vspace{-7pt}
\caption[]{Examples of camera pose estimation results of our approach on 10 different objects from both the synthetic NeRF (1st row) and the Tanks and Temples (2nd row) datasets. Green indicates the ground-truth camera and red indicates the estimated camera.}
\vspace{-8pt}
\label{fig:qualitative_results}
\end{figure*}

\subsection{Discussion on results}
We report the quantitative results in Tab.~\ref{tab:per_object_results} and the qualitative results in Fig.~\ref{fig:qualitative_results} for both datasets.

\noindent\textbf{Effect of initialization on iNeRF:} In both Synthetic NeRF (Tab.~\ref{tab:per_object_results}a) and Tanks and Temples (Tab.~\ref{tab:per_object_results}b), iNeRF performs best when initialized from poses near the known camera (following \cite{yen2020inerf} evaluation). 
However, in more realistic settings with random initialization, the relative error is $21.2^{\circ}$ degrees and $0.218$ units worse in terms of angular and translation errors, respectively.

\noindent\textbf{Synthetic \& real datasets:}
Tab.~\ref{tab:per_object_results}a shows that our \NETNAME consistently outperforms iNeRF with random initialization reducing the average angular error by $72.2^{\circ}$ degrees and the translation by $1.29$ units, respectively, in the case of the fine-tuned backbone.
The same significant performance improvement can be observed in Tanks \& Temples (real).
Tab.~\ref{tab:per_object_results}b shows a $66.5^{\circ}$ and $1.129$ improvement for rotation and translation errors, respectively. 
These results suggest that \NETNAME can be resilient to images taken from a wide range of camera poses.
Conversely, iNeRF with the initialization bounds specified in \cite{yen2020inerf}, on synthetic data, consistently achieves better results.
However, this behavior of iNeRF is not confirmed with real data, whereas \NETNAME clearly stands out with better performance while not requiring constrained initialization.
Furthermore, applying iNeRF as a refinement step for \NETNAMEnoSpace's solution leads to improved results, especially on real data.

\noindent\textbf{Comparison of backbones and noise robustness:}
The effect of pretrained \textit{vs.}~fine-tuned backbones depends on the dataset.
In Synthetic NeRF (Tab.~\ref{tab:per_object_results}a), we can observe a significant improvement with the fine-tuned backbone for both rotation and translation errors.
However, in Tanks \& Temples (Tab.~\ref{tab:per_object_results}b), we can observe that only the translation error slightly decreases by $0.047$, while the rotation error increases by $0.2$.
Such small variation may be caused by the pretrained DINOv2 as it can already extract highly informative features from real-world data.
Our architecture seems also to be robust to noise, without re-training in the NeRF-synthetic dataset, and despite introducing a 50\% Gaussian noise to the image, the errors increased only by $14.01\degree$ and $0.201u$.
Synthetic data might represent a domain shift because it is unlikely that they were used during training.

\noindent\textbf{\NETNAME as initialization to iNeRF:} 
As the output pose of \NETNAME provides a near estimate of the solution, we can also use this as an initialization for iNeRF optimization process. 
It can be seen that in Synthetic NeRF, in contrast to our best-performing result, only the mean translation error decreases by $0.078$, whereas the angular error increases by $3.8$. 
In Tanks and Temples, both translation and angular errors decrease by $0.049$ and $0.6$ respectively. 
However, it is interesting to observe that, in contrast to iNeRF with its original initialization \cite{yen2020inerf}, the combination of \NETNAME and iNeRF performs relatively poorly on synthetic (by $15.2$ and $0.399$ higher angular and translation error, respectively), but significantly better on real-world data.
This could be due to the fact that real data present shadows in the scenes.
The camera and its handling system is not transparent, as in the synthetic data.
These generated shadows seem to affect iNeRF, especially in the case of lateral translation.

\noindent\textbf{Computational performance:}
In terms of inference time, \NETNAME is significantly more efficient than iNeRF, achieving real-time performance at 34fps.
In the case of \NETNAME with iNeRF optimization, we revert to the running time of iNeRF; however, as noted in the case of real-world datasets (\textit{i.e.} Tanks \& Temples) this step further improves accuracy.
In addition, it is worth mentioning that our method is also faster than the CROSSFIRE~\cite{moreau2023crossfire} and Parallel-iNeRF~\cite{lin2023parallelinerf}, where on the same hardware they report $200$ms and $20$s, respectively, compared to $29$ms of our solution.
A Limitation of \NETNAME is to require training for each scene.
Finally, in terms of memory, at test time \NETNAME only uses 3294 MB of memory as opposed to the 4870 MB of iNeRF.

\section{Conclusions}\label{sec:conclusion}

We have proposed a ray sampling by attention method for estimating 6DoF camera poses given a single image and a NeRF model of the scene.
Our experimental evaluation shows that \NETNAME can achieve surprising results without the need for an initialization while being faster and requiring fewer memory resources.
The method advantages stem from the ray generation strategy that can efficiently sample a wide range of camera poses hypothesis coupled with the efficiency of the attention module that maps rays to image pixel features.
The proposed method achieves improved robustness over both synthetic and real-world datasets, while it can also be adopted for real-time applications in robotics and other fields.
Future research will be focused on improving the accuracy between the registered model and the observed target image together with generalising the model to account for several scenes and objects.

{
    \small
    \bibliographystyle{IEEEtran}
    \bibliography{main}
}

\end{document}